# ONLINE INTERPRETATION OF NUMERIC SIGN LANGUAGE USING 2-D SKELETAL MODEL

S. Basu, S. Dey, K. Mukherjee, T. S. Jana
Computer Sc. & Engg. Dept., MCKV Institute of Engineering,
243,G T Road (North); Liluah, Howrah-711204; India.
Email: subhadip8@yahoo.com

*Abstract* - Gesturing is one of the natural modes of human communication. Signs produced by gestures can have a basic meaning coupled with additional information that is layered over the basic meaning of the sign. Sign language is an important example of communicative gestures that are highly structured and well accepted across the world as a communication medium for deaf and dumb. In this paper, an online recognition scheme is proposed to interpret the standard numeric sign language comprising of 10 basic hand symbols. A web camera is used to capture the real time hand movements as input to the system. The basic meaning of the hand gesture is extracted from the input data frame by analysing the shape of the hand, considering its orientation, movement and location to be fixed. The input hand shape is processed to identify the palm structure, fingertips and their relative positions and the presence of the extended thumb. A 2-dimensional skeletal model is generated from the acquired shape information to represent and subsequently interpret the basic meaning of the hand gesture.

## I. INTRODUCTION

Gesture recognition is a user-friendly and intuitive means for man-machine-communication. Numeric sign language is a significant example of structured and globally accepted communicative gestures for deaf and dumb people [6]. Most of the work on vision-based human computer interaction focuses on the recognition of static hand gestures or postures [1],[5]. Wren *et. al.* in [4] track the movement of the human body in real time using 2-d modelling. In [3] a recognition technique is discussed using 2-d Gabor wavelets and subsequent amplitude thresholding and background suppression of familiar objects in an unrestricted environment.

In this paper, a simple and fast technique is used to model numeric hand signs into a meaningful 2-d skeletal model. 2-d skeletal representation is a simple appearance-based model [2] for representation of standard anatomical structure of our hand.

## II. DESCRIPTION OF THE TOOLS AND TECHNIQUES:

### A. Sign Vocabulary

A hand gesture vocabulary [6] with 10 different numeric hand sign representations (fig. 1) is considered in this paper. Only right hand gestures against a contrasting, uniform background is assumed to concentrate the focus on the hand-shape recognition process. Hand gestures are assumed to be upright and completely contained within the input image frame.

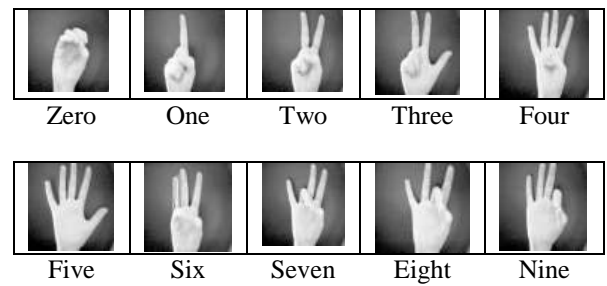

Figure 1. Hand Shapes representing different numeric symbols

### B. System Architecture

Figure 2 illustrates the architecture of the online numeric sign language recognition system. In the first step, the input image is captured through a web camera as a 24-bit bitmap format. In the second step, the bitmap image is converted into a grey level image. In the third step, hand shape contour is detected from the grey level image using Sobel's operator. Analysis of the hand shape contour generates a 2-d skeletal model of the hand sign in the next step. Finally, the input hand gesture is either recognized (classified as a known hand sign) or rejected (classified as an unknown hand sign) according to certain decision rules.

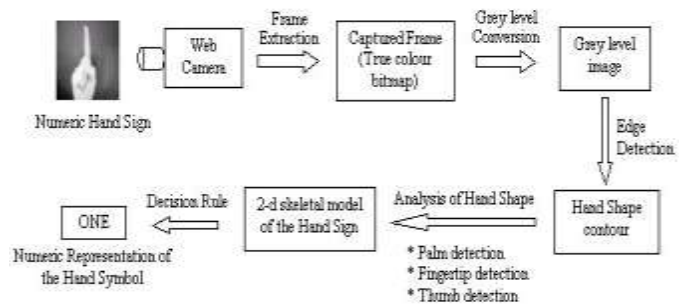

Figure 2. System Architecture

*C. Web Camera Interfacing and Frame Extraction*

Online hand movements are captured using a Logitech QuickCam® Express camera. The Logitech QuickCam® Software Development Kit Version-2 [7], provides necessary tools to develop low-level camera interfaces to capture real-time images using Visual Basic 6. This tool provides a windowed user-interface with live video preview and access to real-time video data to capture 320x240, 24-bit colour image sequences at frame rates of 10 frames per second.

*D. Grey Level Conversion and Edge Detection*

The captured image sequences are stored as windows bitmap format. The brightness component Y of each pixel, which is equivalent to its grey level, can be obtained from the red, green and blue components of the pixel using the following formula;

$$Y = 0.3 * R + 0.59 * G + 0.11 * B$$

The grey level value of each pixel is further analysed using Sobel's operator to detect the edges in the hand shape. If $g_i$ is the grey level value of an image pixel at a given co-ordinate (r,c), then the gradient g'(r,c) is computed, as shown in table 1(b).

| g2 | G1 | g8 |
|----|----|----|
| g3 | **G0** | g7 |
| g4 | G5 | g6 |

Table 1 a. Comparison of grey levels of 8-connected neighbours of g0

| D1=1/4*[(g4+2g5+g6)-(g2+2g1+g8)] |
|---|
| d2=1/4*[(g8+2g7+g6)-(g2+2g3+g4)] |
| g' (r , c) = sqrt[1/2{(d1*d1)+(d2*d2)}] |

Table 1 b. Computation of d1, d2 and gradient g'(r,c)

If the value of the gradient, g', is greater than a pre-defined threshold value, then the pixel at co-ordinate (r , c) is identified as an edge pixel. Figure 3 shows the hand shape contour obtained after the execution of edge detection process on a particular hand shape.

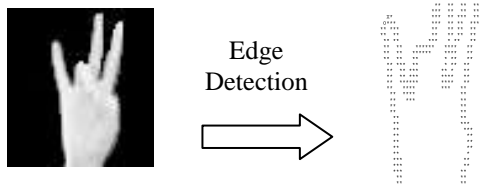

Figure 3. Output of edge detection process

*E. Analysis of Hand Shape*

Analysis of hand shape comprises of the following intermediate processes.

*Palm Detection*

Identification of the palm is the most basic step for the analysis of hand shape. An approximate rectangular boundary is identified for modelling the palm position within the hand shape. Vertical and horizontal histograms representing edge pixel intensity are used with suitable thresholds to identify left, right, top and bottom boundaries of the rectangular palm model.

*Fingertip Detection*

Fingertip positions are identified from the hand shape contours using five simple 2x2 operators T1, T2, T3, T4 and T5 (fig. 4 a). A specific arrangement of these operators signifies the presence of a fingertip in the hand shape. Recognition of each fingertip comprises of 5 intermediate positions or states. A state diagram (fig. 4 b) represents the state transitions depending on a match of input operators. The starting state $s_0$ is reached when a match with the T1 operator is found and the final state $s_4$ signifies successful recognition of the fingertip. Figure 4 c represents a sample hand shape contour and its corresponding fingertip(s).

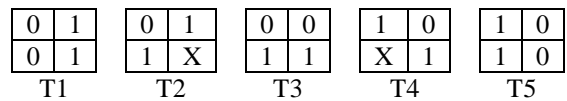

Figure 4 a. Fingertip matching operators

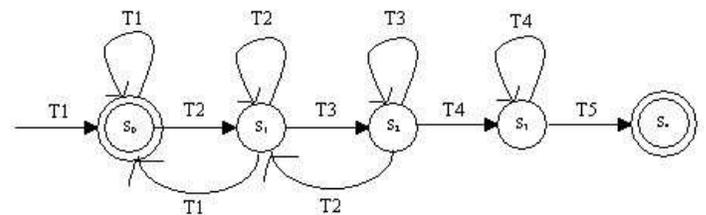

Figure 4 b. State transition diagram on match of input operators

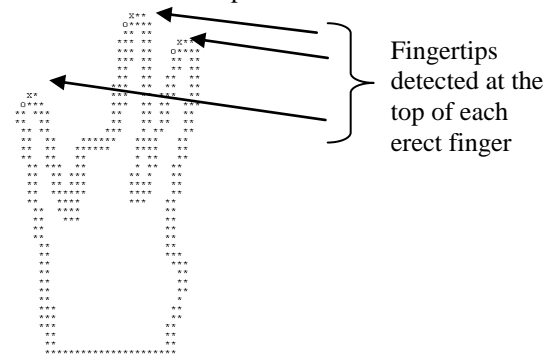

Fingertips detected at the top of each erect finger

Figure 4 c. Fingertips detected on the contour of the hand shape

*Thumb Detection*

Among the ten basic numeric hand symbols, only two representations (three and five) contain an extended thumb sign. In all other representations the ideal location of the thumb is within the palm region and does not contribute any additional information other than those represented by the four fingers. Vertical pixel density histogram of the edge pixels at the right side of the rightmost palm boundary is computed and compared with a predefined threshold to determine the presence of an extended thumb.

*F. Decision Rule*

The scope of gestured interface for Human Computer Interaction is directly related to the proper modelling of the hand gestures. For this application a simple 2-dimensional skeletal model is proposed to interpret the basic meaning associated with the hand gesture. On the top boundary of the palm, four potential finger joints are identified at approximately equal distances. Identified fingertips are connected to the nearest finger joint using a simple finger-tracking technique (fig. 5). Finger lengths less than a predefined threshold are rejected as bent fingers. Final decisions are based on the finger count, their relative connectivity with the finger joints and the presence of extended thumb.

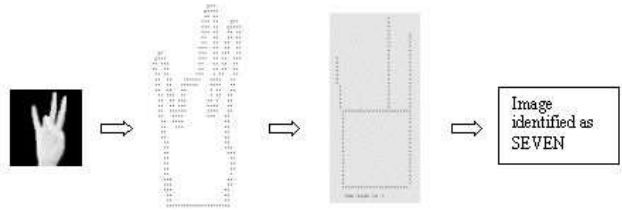

Figure 5. 2-d hand shape modelling from the input image

### III. END RESULTS AND THEIR IMPLEMENTATIONS

The proposed technique is tested with hand gestures of limited samples comprising of 15 different persons of varied age group and 360 different hand shape samples. Only right hand gestures against a plain and contrasting background is considered. The hand shape is assumed to be completely contained within the input frame and is positioned in an upright fashion within 1 meter from the web camera. It is also assumed that any specific hand sign need to be static in a specific shape for at least 1 second for successful modelling of the hand sign. For online recognition, 5 frames are captured per second. If the 2-d modelling of all the 5 consecutive frames is identical, then only an inference is drawn about the possible interpretation of the hand gesture. Otherwise the gesture is ignored as an unintentional movement. Table 2 shows the success rate of the system for both valid and invalid symbols in different sample sets. Each sample set comprises of 100 valid hand signs and 20 invalid hand signs of five persons. Overall success rate for identifying a valid hand sign (within the vocabulary) is 81.33% and that for identifying an invalid hand sign (not within the vocabulary) is 83.33%. Work is still being continued for testing the performance of the system with larger dataset.

|  | Success Rate for Hand signs within the vocabulary | Success Rate for Hand signs not within the vocabulary |
|---|---|---|
| Sample Set A | 76.00% | 80.00% |
| Sample Set B | 82.00% | 85.00% |
| Sample Set C | 86.00% | 85.00% |
| Overall | 81.33% | 83.33% |

Table 2. Performance of the system

Sample output for hand signs within the vocabulary (fig. 6a and fig. 6b) and output for hand signs not within the vocabulary (fig. 6 c) are shown below.

| Hand sign image | Hand shape contour | 2-d skeletal model |
|---|---|---|
|  |  |  |
|  |  |  |
|  |  |  |
|  |  |  |
|  |  |  |

Figure 6 a . 2-d hand shape modelling of the hand signs within the vocabulary

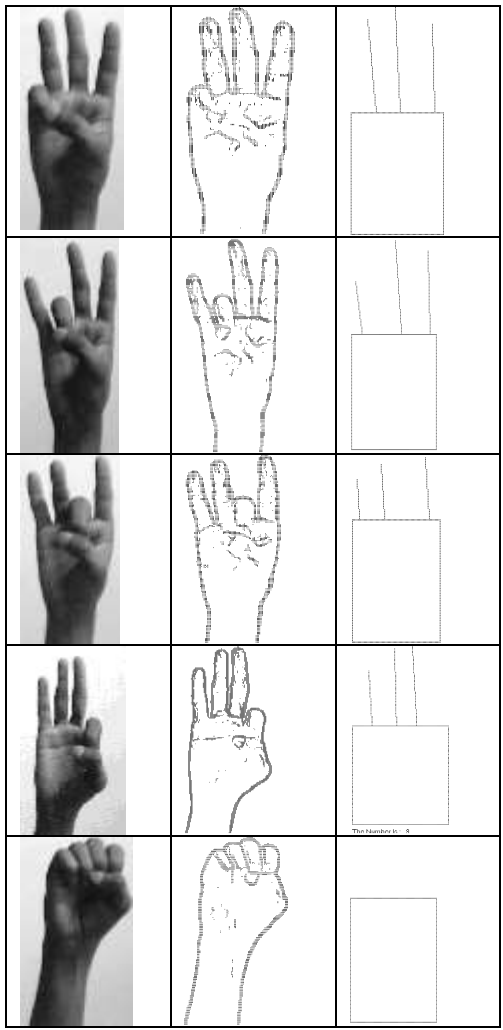

Figure 6 b . 2-d hand shape modelling of the hand signs within the vocabulary (contd.)

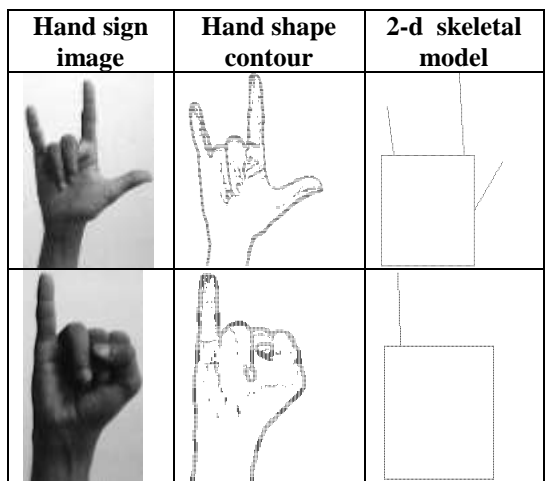

Figure 6 c . 2-d hand shape modelling of the hand signs not within the vocabulary

Human Computer Interaction is having an immense importance in the area of interpretation of sign language for deaf and dumb. Using this technique, basic meaning of all numeric hand gestures is interpreted successfully in restricted environment. Lot of scope for future work exists in the area of interpretation of alphabetic sign languages for deaf and dumb. The present technique can be further enriched by relaxing some of the environmental restrictions and introducing more detailed models of the hand shapes.

ACKNOWLEDGEMENT

Authors are thankful to the "Project on Storage Retrieval and Understanding of Video for Multimedia" , "Centre for Microprocessor Application for Training Education and Research", CSE Department, Jadavpur University,  and MCKV Institute of Engineering, Liluah, Howrah, for providing necessary infrastructure facilities during the progress of the work. Authors are also thankful to Mr. Nitin Jain, Mrs. Arati Gupta, Mr. Kankan Mondal, Mr. Anupam Mondal, Mr. Debasis Mondal and Mr. Debojyoti Mallick for their invaluable contribution towards the successful completion of the project.

REFERENCES

[1] Ong SCW, Ranganath S, Venkatesh YV, "Deciphering Layered Meaning in Gestures", *Proc. of 16th ICPR (III:815-818),* Aug-2002, Canada.
[2] V.I. Pavlovic, R. Sharma, T. S. Huang, "Visual Interpretation of Hand Gestures for Human-Computer Interaction: A Review", *IEEE Trans. Pattern Analysis and Machine Intelligence,* vol. 19, no.7, pp. 677-694, July-1997.
[3] R.P.Würtz, "Object Recognition Robust Under Translations, Deformations, and Changes in Background",  *IEEE Trans. Pattern Analysis and Machine Intelligence,* vol. 19, no. 7, pp. 769-775, July-1997.
[4] C.R.Wren, A. Azarbayejani, T. Darrell, A.P. Pentland, "Pfinder: Real-Time Tracking of the Human Body", *IEEE Trans. Pattern Analysis and Machine Intelligence,* vol. 19, no. 7, pp. 780-785, July-1997.
[5] C. Nölker, N. Ritter, "GREFIT: Visual Recognition of Hand Postures",*http://www.techfak.uni_bielefeld.de/ags/ni/projects/humancompinterface/vishand_e.htm.*
[6]  *www.handspeak.com*
[7] Logitech QuickCam® Software Development Kit Version-2, *www.logitech.com*